\documentclass[conference]{IEEEtran}
\IEEEoverridecommandlockouts
\usepackage{cite}
\usepackage{amsmath,amssymb,amsfonts}
\usepackage{algorithmic}
\usepackage{graphicx}
\usepackage{textcomp}
\usepackage{xcolor}
\def\BibTeX{{\rm B\kern-.05em{\sc i\kern-.025em b}\kern-.08em
    T\kern-.1667em\lower.7ex\hbox{E}\kern-.125emX}}
\usepackage[hyphens]{url}
\usepackage{hyperref}
\usepackage[hyphenbreaks]{breakurl}
\usepackage{booktabs}
\usepackage{multirow}

\usepackage{CJKutf8}
\usepackage{xcolor}
\def\BibTeX{{\rm B\kern-.05em{\sc i\kern-.025em b}\kern-.08em
    T\kern-.1667em\lower.7ex\hbox{E}\kern-.125emX}}
\usepackage[normalem]{ulem}
\usepackage{ascmac}
\usepackage{tcolorbox}
\tcbuselibrary{raster,skins,breakable}
\tcbuselibrary{xparse} 
\DeclareTColorBox{brekableitembox}{ o m O{.5} O{} }%
  {empty, left=2mm, right=2mm, top=-2mm, bottom=0.2mm, attach boxed title to top left={xshift=1.2em},
  boxed title style={empty,left=-2mm,right=-2mm}, colframe=black, coltitle=black, coltext=black, breakable,  
  underlay unbroken={\draw[black,line width=#3pt]
    (title.east) -- (title.east-|frame.east) -- (frame.south east) -- (frame.south west) -- (title.west-|frame.west) -- (title.west); },
  underlay first={\draw[black,line width=#3pt](title.east) -- (title.east-|frame.east) -- (frame.south east) ;
    \draw[black,line width=#3pt] (frame.south west) -- (title.west-|frame.west) -- (title.west); },
  underlay middle={\draw[black,line width=#3pt](frame.north east) -- (frame.south east) ;
    \draw[black,line width=#3pt](frame.south west) -- (frame.north west) ;},
  underlay last={\draw[black,line width=#3pt](frame.north east) -- (frame.south east) -- (frame.south west) -- (frame.north west) ;}, IfValueTF={#1}{title=~~#2~~〈#1〉}{title=~~#2~~},#4}

\newcommand*{\Ja}[1]{%
  \begin{CJK}{UTF8}{ipxm}#1\end{CJK}}

\begin{document}
\title{
Refined and Segmented Price Sentiment Indices\\from Survey Comments
}

\author{\IEEEauthorblockN{Masahiro Suzuki}
\IEEEauthorblockA{
\textit{Nikko Asset Management Co., Ltd. /}\\\textit{The University of Tokyo}\\
Tokyo, Japan \\
research@msuzuki.me
}\and
\IEEEauthorblockN{Hiroki Sakaji}
\IEEEauthorblockA{
\textit{Hokkaido University}\\
Hokkaido, Japan \\
sakaji@ist.hokudai.ac.jp
}
}

\maketitle

\begin{abstract}
We aim to enhance a price sentiment index and to more precisely understand price trends from the perspective of not only consumers but also businesses.
We extract comments related to prices from the Economy Watchers Survey conducted by the Cabinet Office of Japan and classify price trends using a large language model (LLM).
We classify whether the survey sample reflects the perspective of consumers or businesses, and whether the comments pertain to goods or services by utilizing information on the fields of comments and the industries of respondents included in the Economy Watchers Survey.
From these classified price-related comments, we construct price sentiment indices not only for a general purpose but also for more specific objectives by combining perspectives on consumers and prices, as well as goods and services.
It becomes possible to achieve a more accurate classification of price directions by employing a LLM for classification.
Furthermore, integrating the outputs of multiple LLMs suggests the potential for the better performance of the classification.
The use of more accurately classified comments allows for the construction of an index with a higher correlation to existing indices than previous studies.
We demonstrate that the correlation of the price index for consumers, which has a larger sample size, is further enhanced by selecting comments for aggregation based on the industry of the survey respondents.
\end{abstract}


\section{Introduction}
Capturing price trends is important in the formulation of economic policies and corporate management strategies.
In Japan, the Consumer Price Index (CPI) is widely used as data to capture price trends.
The CPI measures the time-series changes in prices by aggregating the prices of goods and services purchased by general consumers.
It plays a crucial role in the formulation of economic policies, wage negotiations, and pension adjustments, and is also widely used as an indicator to assess the health of the economy.
Given that the CPI is particularly important for inflation forecasting, its prediction has become a significant subject of research~\cite{nakajima2023estimation, BARKAN20231145, JOSEPH20241521}.

In recent analyses of CPI and inflation expectations, text data is increasingly being used in addition to traditional numerical data from statistical surveys~\cite{guzman2011internet,seabold2015nowcasting}.
In Japan, the Bank of Japan, which serves as the central bank, conducts research on inflation expectations using the Economy Watchers Survey~\cite{otaka2018economic,nakajima2021extracting}.
However, previous studies have primarily relied on word-based analysis and have not fully leveraged the benefits of language models, which have rapidly advanced in recent years.

Large language models (LLMs), such as ChatGPT and GPT-4, have demonstrated high performance in natural language processing tasks.
The application of LLMs is not limited to general domains but is also expanding into specialized fields such as medicine~\cite{Sukeda-2024-aih}, law~\cite{jayakumar-etal-2023-legal}, and finance~\cite{tanabe2024jafin}.
Unlike traditional models that require fine-tuning, LLMs exhibit different behaviors, such as solving tasks in a zero-/few-shot manner and answering questions based on instructions~\cite{brown2020language,kojima2022large}.
Furthermore, these models can solve tasks across various languages and domains with high performance using a single model without specialized training in multiple languages or domains.
It has become possible to extract richer information from text data with the evolution of language models.
Therefore, utilizing the capability of the language models is expected to further enhance inflation forecasts.

Previous research on inflation expectations has primarily focused on the CPI, and the potential for analysis concerning the corporate price index has not yet been fully explored.
In Japan, as shown in Table~\ref{tab:rel-price-index}, there are price indices for goods and services, as well as for consumers and corporations.
For consumer price indices in Japan, the Ministry of Internal Affairs and Communications conducts segmented price surveys for each item, which are then aggregated based on various criteria such as goods and services.
The comprehensive CPI, which combines consumer goods and services, is commonly the subject of research in Japan and is often referred to as the core core CPI\footnote{This index is calculated from the all items, excluding fresh food, which is affected by weather, and energy, which is susceptible to oil prices.}.
In Table~\ref{tab:rel-price-index}, the core core CPI is represented by the integration of CPI (Goods) and CPI (Services).
The corporate price index indicates price trends in inter-company transactions.
In Japan, the Bank of Japan publishes the Corporate Goods Price Index (CGPI) and the Services Producer Price Index (SPPI) for goods and services, respectively.
Previous research reported that the corporate price index has a leading correlation with the CPI~\cite{akcay2011causal,SASAKI2022102599,sun2023dynamic}.
However, most existing studies focus only on the CPI, and there are few that specifically focus on the corporate price index~\cite{SASAKI2022102599,nakajima2021extracting}.
The reason is considered to be that the corporate price indices reflects the prices of transactions between companies, and these data are seldom publicly available, making data collection difficult.
Therefore, it would be possible to achieve more accurate inflation forecasts by focusing not only to consumer price indices but also to corporate price indices.

In this study, we construct new price sentiment indices (PSIs) using text information from the Economy Watchers Survey and language models.
We aim to refine inflation forecasts with LLMs more accurately than traditional word-based methods.
We demonstrate the potential to enhance classification performance by integrating outputs from multiple LLMs.
In addition to the PSI aimed at the core core CPI, which has been the main focus of previous research, we newly construct four types of PSIs with characteristics as shown in Table~\ref{tab:rel-price-index}, which are combinations of consumer and corporate, goods and services.
We propose a method that extracts only comments relevant to each purpose by using information on the domains and industries of respondents included in the Economy Watchers Survey to construct the PSIs.
Our analysis shows that our segmented PSIs have higher correlations with the three consumer price indices and the comprehensive PSI has higher correlations with the CGPI and the SPPI than the indices by the comparative method.
Our contributions are as follows:

\begin{itemize}
  \item We propose a framework that integrates multiple LLMs and develops more segmented PSIs from comments classified from the perspectives of  consumer and business, goods and services
  \item We demonstrate that text classification related to prices can be performed better with LLMs and that integrating multiple LLMs can potentially yield better performance
  \item The more segmented PSIs by our framework showed the highest correlation with the three consumer-related price indices compared to the previous studies, highlighting the importance of classifying  comments in more detail
  \item Our comprehensive PSI showed leading correlations with existing CGPI and SPPI, indicating its potential usefulness in predicting corporate price indices
\end{itemize}

\begin{table}[tbp]
\caption{
The index related to prices in Japan.
The combination of CPI (Goods) and CPI (Services) is generally referred to as (core core) CPI.
}
\label{tab:rel-price-index}
\begin{center}
\begin{tabular}{ccc} \toprule
     & Consumer & Corporate \\ \midrule
    Goods & CPI (Goods) & CGPI \\
    Services & CPI (Services) & SPPI \\
    \bottomrule
\end{tabular}
\end{center}
\end{table}

\section{Related Works}
\subsection{Price Index Forecast}
Price indices have garnered significant attention as a subject of research, with particular focus on the consumer price index, where regression analysis and deep learning-based predictions are frequently conducted~\cite{BARKAN20231145, nguyen2023consumer, JOSEPH20241521,rohmah2021comparison}.
As a prediction method using text, Chakraborty et al. extracted events from news articles and predicted fluctuations in crop prices based on the event information~\cite{Chakraborty2016}.
Their event-driven prediction model demonstrated that incorporating event information into a standard ARIMA model reduced the RMSE by 22\%.
Shapiro et al. expanded the financial vocabulary to build a more accurate sentiment analysis model and extracted sentiment scores from the text information of newspaper articles related to economics and finance~\cite{SHAPIRO2022221}.
By calculating the impulse response to sentiment shocks, they discovered that positive sentiment shocks suppress inflation.
Angelico et al. used tweets in Italy to construct an indicator of consumer inflation expectations~\cite{ANGELICO2022259}.
Their daily indicator showed a high correlation with both monthly survey-based and daily market-based inflation expectations, and further demonstrated a leading tendency.

\begin{figure*}[tbp]
\centerline{\includegraphics[keepaspectratio,width=\linewidth]{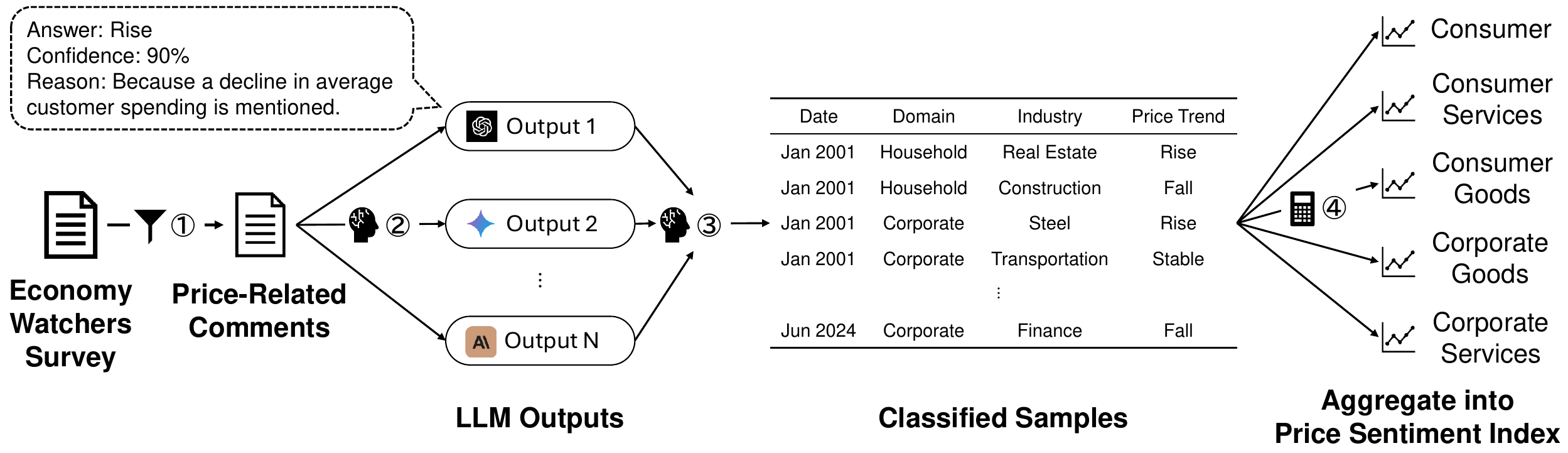}}
\caption{Overview processes of aggregating indicies from the Economy Watchers Survey.
(1) Extract comments related to prices from the Economy Watchers Survey.
(2) Using multiple LLMs, classify the extracted comments to determine whether they refer to increases, stability, or decreases in prices, and output the classification results along with the confidence level and reasons.
(3) Integrate the outputs of multiple LLMs using an LLM to make a final classification regarding the direction of prices indicated by the comments.
(4) Aggregate the classified comments to the five price sentiment indices monthly using the information on the domain and the respondent's industry.
}
\label{fig:overview}
\end{figure*}

\subsection{Economy Watchers Survey}
The Economy Watchers Survey is a monthly survey published by the Cabinet Office, one of central government ministries in Japan, since January 2000.
It gathers responses based on subjective evaluations from approximately 1,800 individuals who are sensitive to economic conditions.
The survey aims to accurately and promptly grasp economic trends and serve as foundational material for economic trend assessments.
The Economy Watchers Survey includes various information such as evaluations and comments on economic trends, reasons for evaluations, and labels related to the fields of the comments.
Some studies using the Economy Watchers Survey have focused on learning the text and sentiments to construct models that assign sentiments to different texts~\cite{yamamoto2016jsai,SEKI2022102795,suzuki2024jsai,goshima2021forecasting}.
Goshima et al. applied a sentiment model trained on the Economy Watchers Survey to newspaper articles to construct an economic trend index~\cite{goshima2021forecasting}.
Seki et al. applied a sentiment model constructed in the same manner as \cite{goshima2021forecasting} to newspaper articles and developed a business sentiment indicator called S-APIR~\cite{SEKI2022102795}.
Suzuki applied the sentiment model to the Japanese Company Handbook, a text database of listed companies in Japan, and demonstrated that the sentiment model's scores are effective in gaining returns for smaller-scale stocks~\cite{suzuki2024jsai}.

Several studies have analyzed the relationship between the Economy Watchers Survey and a price index~\cite{otaka2018economic,nakajima2021extracting,nakajima2022extracting}.
Otaka and Kan utilized co-occurrence networks to construct a price sentiment index, which they found to be useful as a leading indicator of the inflation rate~\cite{otaka2018economic}.
They also extracted words that have explanatory power for expected inflation rates and actual inflation rates based on the variable importance of random forests.
Nakajima et al. extended the research~\cite{otaka2018economic} by constructing an index using only a limited number of words~\cite{nakajima2021extracting}.
They demonstrated that the constructed index captures cost factors such as fluctuations in raw material prices and exchange rates.
Nakajima et al. developed an indicator that quantitatively represents firms' short-term inflation expectations~\cite{nakajima2022extracting}.
This indicator tends to lead consumer price inflation by several months and has been shown to move in conjunction with demand and cost variables.

\subsection{Large Language Models in Finance}
Regarding the performance of LLMs when applied to the financial domain, while they demonstrate high performance across many financial domain tasks, fine-tuned models using BERT, financial domain-adapted BERT (FinBERT), or RoBERTa tend to perform better when proprietary datasets are used~\cite{guo-etal-2023-chatgpt, li-etal-2023-chatgpt}.
LLMs have the ability to generate text, which has led to attempts to extend conventional financial tasks or create new tasks to verify the performance of LLMs in the financial domain~\cite{xie2024finben,hirano-2024-construction}.
There are also efforts to utilize ChatGPT for portfolio construction and stock selection~\cite{romanko2023chatgpt,KO2024105433,PELSTER2024104786}.
\cite{wang-etal-2024-llmfactor,Koa2024www} used LLMs to explain stock price predictions, providing not only forecasts but also the reasoning and background knowledge.
Research on the effectiveness and utilization of LLMs in the financial and economic domains is currently in the midst of development.

\section{Methods}
Figure \ref{fig:overview} illustrates the overview of our proposed method to construct price sentiment indices (PSIs).
The process of creating our proposed PSIs is divided into four steps.
First, we extract only the comments related to prices from the Economy Watchers Survey (Figure \ref{fig:overview}, (1)).
For the extracted comments, we classify whether they refer to increases, stability, or decreases in prices. using LLM, and output the results along with confidence levels and reasons (Figure \ref{fig:overview}, (2)).
The outputs obtained from multiple LLMs for each sample are integrated by the LLM to produce a single direction of price movement (Figure \ref{fig:overview}, (3)).
We create respective indices of consumer prices, consumer goods prices, consumer service prices, corporate goods prices, and corporate service prices by aggregating the obtained results (Figure \ref{fig:overview}, (4)).

\subsection{Price-related Texts Filtration}\label{sec:filtering}
We extract comments related to prices from the Economy Watchers Survey (Figure \ref{fig:overview}, (1)).
Given that nearly 2,000 comments are published each month, inputting all comments from the Economy Watchers Survey into multiple LLMs is not cost-effective. 
Therefore, we extract comments related to prices and then classify the direction of prices.

We first create labeled data to verify the performance of the filtration.
We randomly extract 1,000 comments that discuss household trends and corporate trends from the Economy Watchers Survey conducted between January 2000 and December 2009.
For the 1,000 extracted comments, the authors manually label whether they are related to prices, referencing the samples described in the previous literature~\cite{otaka2018economic}.
As a result of the annotation, 304 samples are identified as related to prices, and 696 samples are identified as not related to prices.
The data used for the Economy Watchers Survey is the 2024.07.0 version (data up to June 2024) developed in \cite{suzuki2024ews}.
The Economy Watchers Survey includes two survey items: the current economic conditions and the future economic outlook.
Previous studies have reported that the PSI constructed from both the current and future surveys exhibited similar trends~\cite{otaka2018economic,nakajima2021extracting}.
Therefore, we utilize only the survey data concerning the current economic conditions.
Of the labeled data, 70\%, 10\%, and 20\% are allocated as train, dev, and test data, respectively.

The experiment uses the following three types of models: fune-tuned models, LLMs, Naive Bayes.

\subsubsection*{Fine-tuned Models}
As Japanese language models for fine-tuning, we employ two models: DeBERTaV2~\cite{Suzuki-2024-findebertav2} as a general-purpose model and FinBERT~\cite{Suzuki-2023-ipm} as a domain-specific model.

\subsubsection*{LLMs}
As a model utilizing LLM, an evaluation is conducted on two models each provided by APIs from OpenAI, Anthropic, and Google.
Five samples are commonly selected from the train and dev data to perform 5-shots in-context learning.

\subsubsection*{Naive Bayes}
We adopt the Naive Bayes classifier employed in \cite{nakajima2021extracting} as a comparative method.
For the 20 words selected in \cite{nakajima2021extracting}\footnote{The specific words are listed in footnote 7 of \cite{nakajima2021extracting}.}, we estimate the scores of each word using the train data and apply them to the test data.
For word segmentation, we use the A mode of Sudachi~\cite{takaoka-etal-2018-sudachi}.

The results are shown in Table~\ref{tab:result-is-related}.
The fine-tuned models, FinBERT and DeBERTaV2, exhibit high performance, with FinBERT achieving the highest performance.
The high performance of the fine-tuned models is considered to be due to the large sample size and the relatively low difficulty of the task.
The LLM also showed higher performance compared to the Naive Bayes, with GPT-4o being only slightly inferior to the fine-tuned models.

We classify all data from the Economy Watchers Survey regarding household and corporate trends using the fine-tuned FinBERT, which demonstrated the highest performance.
Since the number of respondents in the survey's starting year (2000) is significantly lower compared to subsequent years, we use data from 2001 onwards.
Out of the total 280,825 entries, 67,060 entries are extracted as being related to prices.

\begin{table}[tbp]
\caption{
Results of filtering price-related texts.
Bold and underline indicate the first and second highest performance, respectively.
GPT, Claude, and Gemini all utilize in-context learning with 5 shots.
}
\label{tab:result-is-related}
\begin{center}
\begin{tabular}{lc} \toprule
    Model & Weighted F1 \\ \midrule
    Naive Bayes~\cite{nakajima2021extracting} & 81.3 \\
    FinBERT & \textbf{96.0} \\
    DeBERTaV2 & \underline{95.5} \\
    GPT-4o-mini & 89.3 \\
    GPT-4o & 94.4 \\
    Claude 3 Haiku & 77.9 \\
    Claude 3.5 Sonnet & 90.0 \\
    Gemini 1.5 Flash & 86.0 \\
    Gemini 1.5 Pro & 90.7 \\
    \bottomrule
\end{tabular}
\end{center}
\end{table}

\subsection{Price Direction Classification}\label{sec:direction-clf}
We classify the comments from the Economy Watchers Survey related to prices, extracted in Section \ref{sec:filtering}, to classify the direction of price changes (Figure \ref{fig:overview}, (2)).
We label two datasets: 304 comments related to prices labeled in Section \ref{sec:filtering} (referred to as Price Direction Data 1) and 1,000 comments extracted using a fine-tuned FinBERT model constructed in Section \ref{sec:filtering} (referred to as Price Direction Data 2).
These datasets are combined for analysis.
Among the 304 comments in Price Direction Data 1, 87 indicate a rise in prices, 40 indicate stability, and 177 indicate a decline.
For Price Direction Data 2, 1,000 comments are randomly selected from the Economy Watchers Survey comments on household and corporate trends from January 2000 to December 2009, which are classified as related to prices by the fine-tuned FinBERT model, ensuring no overlap with the data extracted in Section \ref{sec:filtering}.
Of these 1,000 extracted comments, 126 indicate a rise, 89 indicate stability, 683 indicate a decline, and 102 are unrelated to prices.

We execute the classification task for the total 1,304 samples combined from Price Direction Data 1 and Price Direction Data 2, similar to the process described in Section \ref{sec:filtering}.
In the classification using LLM, not only the classification results but also the confidence level and the reason are output to integrate the outputs of multiple models in subsequent steps.
The classification by a LLM is performed through 5-shots in-context learning.

The results of the experiment are shown in Table \ref{tab:class}. The performance of the LLM models is high, with the GPT-4o, Claude 3.5 Sonnet, and Gemini models demonstrating performance that surpasses that of the fine-tuned models.
Due to the task being more challenging than those related to prices discussed in Section \ref{sec:filtering}, it is possible that the larger model size of the LLMs resulted in higher performance even without fine-tuning.
By outputting confidence levels and reasons, the scores of GPT-4o-mini and Claude 3.5 Sonnet slightly decrease, while positive effects are observed in the other models.

\begin{table}[tbp]
\caption{
Results of text classification of price trends.
Bold and underline indicate the first and second highest performance, respectively.
GPT, Claude, and Gemini all utilize in-context learning with 5 shots.
}
\label{tab:class}
\begin{center}
\begin{tabular}{lc} \toprule
    Model & Weighted F1 \\ \midrule
    Naive Bayes~\cite{nakajima2021extracting} & 64.6 \\
    FinBERT & 77.3 \\
    DeBERTaV2 & 80.0 \\
    GPT-4o-mini & 73.8 \\
    GPT-4o & 80.2 \\
    Claude 3 Haiku & 74.9 \\
    Claude 3.5 Sonnet & 82.4 \\
    Gemini 1.5 Flash & 84.7 \\
    Gemini 1.5 Pro & 80.8 \\ \midrule
    \textbf{Model w/ confidence and reason} & \\
    GPT-4o-mini & 72.7 \\
    GPT-4o & \underline{85.0} \\
    Claude 3 Haiku & 77.8 \\
    Claude 3.5 Sonnet & \textbf{85.3} \\
    Gemini 1.5 Flash & 84.4 \\
    Gemini 1.5 Pro & 82.4 \\    
    \bottomrule
\end{tabular}
\end{center}
\end{table}

\subsubsection*{Integration with Multiple Outputs}\label{sec:integrate}
The classification outputs from multiple LLMs are integrated into a single classification output (Figure \ref{fig:overview}, (3)).
Here, we integrate the outputs of three models: GPT-4o from OpenAI, Claude 3.5 Sonnet from Anthropic, and Gemini 1.5 Flash from Google.
The following box shows an example of the prompt.
Although the original text is in Japanese, we provide the English translation version.
The classification results, confidence levels, and reasons output from each model are input into a single model, which then performs classification again based on these inputs.

\begin{brekableitembox}{Integrated Price Direction Classification}
\fontsize{9.5pt}{12pt}\selectfont
Based on the outputs of several models, classify the given text according to the type of price change it refers to. If it mentions a rise, classify it as ``Rise''; if it states that there is no change and prices are stable, classify it as ``Stable''; if it refers to a fall, classify it as ``Fall''; and if it does not mention any price change, classify it as ``Not related.'' Do not provide any output other than these classifications.\\
\\
Text: Due to the decrease in summer visitors, surrounding courses significantly lowered their play fees in September, resulting in a decrease in visitors to our golf course, but the average spending per customer has not dropped significantly.\\
\\
\Ja{【}Model 0\Ja{】}\\
Classification Result: Stable\\
Confidence: 80\%\\
Reason: The text clearly states that ``the average spending per customer has not dropped significantly.'' Despite other golf courses lowering their fees, the average spending per customer at this golf course has not changed significantly, leading to the classification as stable.\\
\\
\Ja{【}Model 1\Ja{】}\\
(Authors' note: Outputs from other models follow and we omit them for brevity.)\\
\\
Classification result considering the above:
\end{brekableitembox}

The classification results after integrating the outputs of these three models are shown in Table \ref{tab:integrate}.
A slightly better performance is achieved compared to using only a single model by integrating the outputs of multiple models.

Due to the issue of API costs, from this point onward, we employ the outputs of GPT-4o and Gemini 1.5 Flash with the integrated classification results from Gemini 1.5 Flash to construct PSIs.

\begin{table}[tbp]
\caption{
Classification results of price direction with integrated outputs from multiple LLMs.
The integration models consolidate the outputs of three models: GPT-4o, Claude 3.5 Sonnet, and Gemini 1.5 Flash, using different models for the integration process.
}
\label{tab:integrate}
\begin{center}
\begin{tabular}{lc} \toprule
    Model & Weighted F1 \\ \midrule
    No Integration (Claude 3.5 Sonnet) & 85.3 \\
    Integrate w/ GPT-4o & 81.7 \\
    Integrate w/ Claude 3.5 Sonnet & 85.1 \\
    Integrate w/ Gemini 1.5 Flash & \textbf{88.0} \\
    \bottomrule
\end{tabular}
\end{center}
\end{table}

\subsection{Aggregation into Each Index}
We aggregate the classification results conducted in Section \ref{sec:integrate} and compile them as PSI (Figure \ref{fig:overview}, (4)).
For each month (survey period), we quantify how much more numerous the comments indicating price rises, $Rise$, are compared to those indicating price falls, $Fall$, similar to the approach in previous research~\cite{nakajima2021extracting}.
Specifically, using the number of comments indicating stable prices, $Stable$, we define the following equation to calculate a PSI.
\begin{equation}\label{eq:psi}
    \mathrm{PSI} = \frac{Rise - Fall}{Rise + Fall + Stable}
\end{equation}

Table~\ref{tab:our-psi} presents the variation of PSI to construct in this study.
The column labeled Target Index represents the existing price indices used for comparison with the constructed PSI.
In this study, in addition to the General PSI, which is constructed by applying equation \eqref{eq:psi} to all comments related to prices extracted and classified from the Economy Watchers Survey up to Section~\ref{sec:integrate}, we also construct five specific indices (Specific PSIs) using only particular comments.
These indices are constructed by varying the extracted comments based on the domain of the comments (household trends or corporate trends) and the industry of the respondents within the Economy Watchers Survey.
While the domains within the Economy Watchers Survey can be classified using rule-based methods, there are numerous types of respondent industries.
Therefore, we perform preprocessing to remove parentheses to classify respondent industries into manufacturing or non-manufacturing, which reduces the number of the types of the respondent industries to 169, and then manually classify them as either manufacturing or non-manufacturing.

The period of the Economy Watchers Survey used for constructing the index is set from January 2001 to June 2024, as described in Section~\ref{sec:filtering}.

{\tabcolsep=2.5pt
\begin{table}[tbp]
\caption{
Overview of the index constructed in this study.
CPI (Goods) refers to the CPI aggregated for goods excluding fresh food.
}
\label{tab:our-psi}
\begin{center}
\begin{tabular}{lccc} \toprule
    PSI & Domain & Industry & Target Index \\ \midrule
    General & \begin{tabular}{c}Household \&\\Corporate\end{tabular} & \begin{tabular}{c}Manufacturing \&\\Non-manufacturing\end{tabular} & - \\
    Consumer General & Household & \begin{tabular}{c}Manufacturing \&\\Non-manufacturing\end{tabular} & Core Core CPI \\
    Consumer Goods & Household & Manufacturing & CPI (Goods) \\
    Consumer Services & Household & Non-manufacturing & CPI (Services) \\
    Corporate Goods & Corporate & Manufacturing & CGPI \\
    Corporate Services & Corporate & Non-manufacturing & SPPI\\
    \bottomrule
\end{tabular}
\end{center}
\end{table}
}

\begin{table*}[tbp]
\caption{
The time-lagged correlation between the constructed PSIs and the existing price indices.
The value shown is when the time-lagged correlation is highest between the PSIs and the price trend index using a lag.
The number in parentheses indicates the months when the time-lagged correlation is highest, with a positive value indicating that the PSI has the highest correlation when it precedes the price index.
The row of the our specific PSI represents the time-lagged correlation with the indices of Consumer General PSI, Consumer Goods PSI, Consumer Services PSI, Corporate Goods PSI, and Corporate Services PSI, in order from the left column.
}
\label{tab:corr}
\begin{center}
\begin{tabular}{lccccc} \toprule
    Model & Core Core CPI & CPI (Goods) & CPI (Services) & CGPI & SPPI \\ \midrule
    Baseline~\cite{nakajima2021extracting} & 0.583 (14) & 0.680 (7) & 0.602 (15) & 0.778 (3) & 0.438 (15) \\
    Our General PSI & 0.635 (14) & 0.723 (7) & 0.635 (16) & \textbf{0.793} (3) & \textbf{0.510} (15) \\
    Our Specific PSI & \textbf{0.645} (14) & \textbf{0.724} (7) & \textbf{0.677} (16) & 0.617 (3) & 0.403 (14) \\
    \bottomrule
\end{tabular}
\end{center}
\end{table*}

\begin{table}[tbp]
\caption{
Results of the Granger causality test.
We test the null hypothesis that ``A does not Granger-cause B.''
$^{***}$ indicates that the null hypothesis of no Granger causality is rejected at the 1\% significance level.
We select 12 months for the maximum lag length.
}
\label{tab:granger}
\begin{center}
\begin{tabular}{cccc} \toprule
    A & B & F value & p value \\ \midrule
    General PSI & Core Core CPI & 42.616$^{***}$ & 0.0000 \\
    Core Core CPI & General PSI & 0.012$^{\ \ \ }$ & 0.6913 \\
    General PSI & CGPI & 37.767$^{***}$ & 0.0000 \\
    CGPI & General PSI & 2.758$^{\ \ \ }$ & 0.0968 \\
    General PSI & SPPI & 30.634$^{***}$ & 0.0000 \\
    SPPI & General PSI & 0.233$^{\ \ \ }$ & 0.6295 \\
    \bottomrule
\end{tabular}
\end{center}
\end{table}

\section{Results and Discussion}
Table \ref{tab:corr} shows the maximum lagged correlation coefficients between the constructed PSIs and CPI (core core, goods, and services), CGPI, and SPPI.
For any of the indices CPI, CGPI, and SPPI, the General PSI constructed has a higher correlation coefficient than that of previous research~\cite{nakajima2021extracting}, suggesting that it can capture price fluctuations with higher accuracy.
From Table \ref{tab:granger}, the Granger causality test also shows significant explanatory power for consumer prices 12 months ahead.
This suggests that the constructed PSIs contain information on price trends for both households and businesses.

When comparing General PSI and Specific PSI, the Specific PSIs focused on household trends exhibit a higher correlation with each CPI index than the General PSI.
Consumer price trends are captured better than the General PSIs by narrowing the comments in the Economy Watchers Survey to household trends.
In particular, the correlation between Specific PSI and CPI (Services) is approximately 0.04 higher than that with the General PSI.
A stronger correlation with CPI (Goods) is achieved by creating indices with a specific focus.

In terms of correlation with CGPI and SPPI, the correlation of Specific PSI, which narrows comments for the index, is lower than that of General PSI.
This is considered due to the scarcity of comments in the corporate trends domain of the Economy Watchers Survey.
Table~\ref{tab:num-comments} shows the average number of comments per month used in the construction of the PSI.
In constructing the Corporate Goods PSI and Corporate Services PSI, the smaller sample size used compared to other PSIs might make it challenging to accurately capture price trends, resulting in a lower correlation.
Therefore, when estimating CGPI and SPPI, it is considered more accurate to use all comments related to prices extracted from the Economy Watchers Survey, as is done with General PSI.
When examining the lag correlation between CPI and SPPI, the maximum value of 0.920 is observed when CPI leads by three months.
CPI is also a viable method for estimating SPPI, though the lead time is reduced.
In the lag correlation between CGPI and SPPI, the maximum value of 0.608 is observed when CGPI leads by eight months, and in the lag correlation between CGPI and CPI, the maximum value of 0.744 is observed when CGPI leads by eleven months.

\begin{table}[tbp]
\caption{The number of monthly comments to construct PSIs}
\label{tab:num-comments}
\begin{center}
\begin{tabular}{lccc} \toprule
    PSI & \# \\ \midrule
    General & 238 \\
    Consumer General & 187 \\
    Consumer Goods & 139 \\
    Consumer Services & 48 \\
    Corporate Goods & 25 \\
    Corporate Services & 25 \\
    \bottomrule
\end{tabular}
\end{center}
\end{table}

\section{Conclusion}
We propose a framework that utilizes LLMs to extract comments related to prices from the Economy Watchers Survey and calculate price sentiment indices.
We extract the direction of prices mentioned in the comments of the Economy Watchers Survey with higher accuracy than previous studies by employing FinBERT to extract price-related comments and integrating multiple LLMs.
In this process, the classification performance improves in most LLMs by outputting not only the classification results but also the confidence level and reasons for the classification.
Furthermore, we classify the comments in more detail to construct specific sentiment indices for consumer, consumer services, consumer goods, corporate goods, and corporate services by using the domain of the comments and the industry information of the respondents from the Economy Watchers Survey.
Through the better classification performance by LLMs and the segmented classification of comments, our constructed specific price sentiment indices showed higher correlation values with three indices related to consumer prices compared to previous studies and indices constructed from all price-related comments in the Economy Watchers Survey.
The two indices related to corporate prices showed the highest correlation with our General PSI.

Future research can address to verify classification accuracy using local models.
While we utilized models from Anthropic, Google, and OpenAI, which provide APIs, an increasing number of Japanese LLMs that can operate locally are being made available~\cite{Suzuki-2023-bigdata,llmjp2024,Fujii:COLM2024}.
More precise classification can be achievable from the integrated outputs with local LLMs.
Analysis of the relationship with economic indicators other than price indices remains a future research topic.
We aim to construct PSIs and compare them with price indices such as the CPI, CGPI, and SPPI.
However, since previous research has reported that the PSI has a high correlation with the Bank of Japan's Tankan diffusion index (DI)~\cite{nakajima2021extracting}, analyzing how our PSI relates to indicators other than price indices could lead to a further understanding of the characteristics inherent in our PSI.

\section*{Notes}
The opinions expressed in this article belong to the authors alone and do not represent the official views of the their affiliated institutions.
Additionally, all possible errors are solely the author’s own.

\section*{Acknowledgment}
This work was supported by JST PRESTO Grant Number JPMJPR2267, Japan.

\bibliographystyle{IEEEtran}
\bibliography{ref}
\appendix
\subsection{Version of API Models}
The versions of the LLM models used in each experiment are as follows: GPT-4o-mini is 2024-07-18, GPT-4o is 2024-08-06, Claude 3 Haiku is 20240307, Claude 3.5 Sonnet is 20240620, and both Gemini 1.5 Flash and Gemini 1.5 Pro are 001.

\subsection{Experimental Settings}
\subsubsection{Fine-tuned models}
Fine-tuned parameters in FinBERT and DeBERTaV2 models are shown in Table~\ref{tab:params}.

\subsubsection{Prompts for LLM}
The following boxes show examples of prompts used for the LLMs in the sections on price-related texts filtration (Section~\ref{sec:filtering}) and price direction classification (Section~\ref{sec:direction-clf}).
We manually develop the confidence level and reasoning used for in-context learning in price direction classification.
Although the original text is in Japanese, we provide the English translation version.

\begin{table}[tbp]
  \begin{center}
  \caption{Hyper-parameters for fine-tuning in our expertiments}
  \label{tab:params}
  \begin{tabular}{lc} \toprule
    Hyper-parameter & Values \\
    \midrule
    Warmup Ratio & 0.1 \\
    Learning Rates & \{1e-5, 2e-5, 5e-5, 1e-4\} \\
    Batch Size & {32, 64} \\
    Maximum Epochs & 10 \\    
    \bottomrule
  \end{tabular}
  \end{center}
\end{table}

\begin{brekableitembox}{Price-related Texts Filtration}
\fontsize{9.5pt}{12pt}\selectfont
Classify whether the given text refers to prices.
If it does, answer ``Yes'', and if it does not, answer ``No.''
Provide no other output.\\
\\
Text: Due to the surge in crude oil prices, there are price revisions from the wholesalers almost every month. Although market prices are stable at a high level, sales are declining due to restrained usage.\\
Answer: Yes\\
(Authors' note: Other examples follow and we omit them for brevity.)\\
\\
Text: The intensification of discount competition within the industry has led to a decline in sales.\\
Answer: 
\end{brekableitembox}

\begin{brekableitembox}{Price Direction Classification}
\fontsize{9.5pt}{12pt}\selectfont
Classify the given text according to the type of price change it refers to.
If it mentions a rise, answer ``Rise''; if it states that there is no change and prices are stable, answer ``Stable''; if it refers to a fall, answer ``Fall''; and if it does not mention any price change, answer ``Not related.''
Follow this with a confidence level and a brief explanation of the reasoning.\\
\\
Text: Sales of women's outerwear for summer have increased as the products were shifted back from low-priced items to the regular price range, leading to increased customer purchases and sales. The number of visitors remains unchanged, but sales are rising.\\
Answer: Rise\\
Confidence: 100\%\\
Reason: The shift from low-priced items to the regular price range indicates a price increase.\\
\\
Text: The local delicacy industry is experiencing favorable trends in orders, production, and sales for both dried and fresh delicacies. As a result, some companies are operating on holidays and overtime, and factory operation rates are improving. Particularly, sales of gift products are strong, with an increase in orders for high-quality, carefully crafted delicious items, even at higher unit prices.\\
Answer: Not related\\
Confidence: 80\%\\
Reason: Although there is mention of unit prices, there is no reference to the trend of price changes.\\
\\
Text: The number of visitors remains unchanged, but the customer unit price is decreasing. This is partly due to the high temperatures, which have led to poor sales of heating products, and sales are below last year's figures.\\
Answer: Fall\\
Confidence: 100\%\\
Reason: The customer unit price is decreasing.\\
\\
Text: As the Golden Week approaches, it is a month centered around family travel, but applications for overseas trips are fewer than usual, and there are many one-night trips domestically, with many inquiries focusing on low-budget and price-conscious options.\\
Answer: Not related\\
Confidence: 60\%\\
Reason: There is no mention of unit price or price fluctuations.\\
\\
Text: The current accommodation unit price is at a level where it cannot go any lower, and ancillary income is also on a recovery trend, returning to levels from three months ago, mainly among local customers, giving an overall impression of price stabilization.\\
Answer: Stable\\
Confidence: 90\%\\
Reason: The accommodation unit price has stopped decreasing and is stable.\\
\\
\\
Text: The intensification of discount competition within the industry has led to a decline in sales.\\
Answer: 
\end{brekableitembox}

\begin{table}[tbp]
\caption{
The classification results of zero-shot LLMs.
Filtering is the classification of whether the comments in the Economy Watchers Survey are related to prices, with the same experimental setup as in Section~\ref{sec:filtering}.
Direction is the classification of the direction of prices indicated by the comments in the Economy Watchers Survey, with the same experimental setup as in Section~\ref{sec:direction-clf}.
The evaluation metric is Weighted F1.
}
\label{tab:appx-result}
\begin{center}
\begin{tabular}{lcc} \toprule
    Model & Filtering & Direction \\ \midrule
    GPT-4o-mini & 86.7 & 65.3\\
    GPT-4o & 86.9 & 71.7 \\
    Claude 3 Haiku & 79.6 & 70.3 \\
    Claude 3.5 Sonnet & 82.4 & 64.9 \\
    Gemini 1.5 Flash & 82.4 & 77.2 \\
    Gemini 1.5 Pro & 90.7 & 77.7 \\
    \bottomrule
\end{tabular}
\end{center}
\end{table}

\subsection{Classification Results of Zero-shot Settings}
The results of the experiments conducted with zero-shot LLMs for Section~\ref{sec:filtering} and Section~\ref{sec:direction-clf} are shown in Table~\ref{tab:appx-result}.
Compared to Table~\ref{tab:result-is-related}, in the price-related comment filtering task, GPT-4o and Claude 3.5 Sonnet show significant performance improvements through in-context learning, whereas Claude 3 Haiku and Gemini 1.5 Pro exhibit either equivalent performance or a decline.
On the other hand, when compared to Table~\ref{tab:class}, all models show performance improvements in the task of classifying the direction of prices, with some models demonstrating particularly notable improvements.
It is considered that the four-class classification task is more challenging than the binary classification filtering task, and the samples provided through in-context learning contributed effectively.

\subsection{Seasonal Adjustment of Indices}
The CPI often uses figures that have been seasonally adjusted to remove fluctuations caused by seasonal factors.
For instance, clothing prices tend to decrease as the change of seasons approaches.
Previous research~\cite{nakajima2021extracting} uses seasonally adjusted CPI, but the seasonally adjusted CPI data is only available from January 2010 onwards, and data prior to this does not exist.
Additionally, since CGPI and SPPI are not seasonally adjusted, to maintain consistency with these indices, we do not use the CPI with seasonal adjustment.
Table~\ref{tab:corr-cpi-adjust} shows the time-lagged correlation between the indices we constructed and both the seasonally adjusted and non-seasonally adjusted indices from January 2010 onwards, when seasonally adjusted indices are available.
Given that there is no significant difference between the values of the seasonally adjusted and non-seasonally adjusted indices, it is considered that the impact of seasonal adjustment on the framework of the indices constructed in this study is minimal.

\begin{table}[tbp]
\caption{
The maximum time-lagged correlation coefficients between the seasonally adjusted and non-seasonally adjusted CPI indices and the Specific PSI (Consumer General PSI, Consumer Goods PSI, Consumer Services PSI) constructed using our proposed method.
}
\label{tab:corr-cpi-adjust}
\begin{center}
\begin{tabular}{lccc} \toprule
     & All & Goods & Services \\ \midrule
    No Adjustment & 0.808 & 0.823 & 0.664 \\
    Adjusted & 0.807 & 0.813 & 0.671 \\
    \bottomrule
\end{tabular}
\end{center}
\end{table}

\end{document}